\documentclass[11pt]{article}

\usepackage[preprint]{acl}

\usepackage{times}
\usepackage{latexsym}
\usepackage[T1]{fontenc}
\usepackage[utf8]{inputenc}
\usepackage{microtype}
\usepackage{inconsolata}
\usepackage{graphicx}
\usepackage{amsmath}
\usepackage{amssymb}
\usepackage{booktabs}
\usepackage{multirow}
\usepackage{xcolor}
\usepackage{subfigure}
\usepackage{colortbl}

\newcommand{\atrain}{\alpha_{\text{train}}}
\newcommand{\aeval}{\alpha_{\text{eval}}}

\title{Style or Content? \\ Evaluating Style Classifiers with Controlled Content Overlap}

\author{
Zhuo Liu \quad
Haozheng Du \quad
Xiangxiang Xu \quad
Hangfeng He \\
University of Rochester \\
\texttt{zhuo.liu@rochester.edu}
}

\begin{document}
\maketitle

\begin{abstract}
Style classifiers can use content cues that correlate with style labels in naturally collected data, yet we lack a systematic way to measure this reliance. We study this problem with a controlled content overlap setup built on parallel Bible translations. Specifically, we define the overlap parameter $\alpha$ as the normalized residual of mutual information between content identity and style label, so that it measures how much content is shared across style classes: from no shared content ($\alpha=0$) to fully shared content ($\alpha=1$).  Cross-overlap evaluation of RoBERTa-based classifiers shows that low-overlap models degrade when content cues are removed, while high-overlap models transfer more robustly. A cross-style content retrieval probe further shows that content becomes less recoverable as $\alpha$ increases, with training dynamics showing this removal occurs gradually. Together, these results suggest that controlled overlap provides a simple diagnostic for separating style learning from content shortcuts.
\end{abstract}

\section{Introduction}
\label{sec:intro}

Style classification aims to identify how a text is written rather than what it is about. It supports a wide range of natural language processing (NLP) applications~\citep{stamatatos2009survey, jin2022deep}, including authorship analysis~\citep{mikros2007investigating}, genre classification, stylistic rewriting~\citep{briakou2021evaluating}, and controllable text generation. Recent classifiers based on pretrained language models achieve strong progress on standard benchmarks~\citep{carlson2018evaluating}, yet high performance alone does not mean that they have learned transferable stylistic patterns~\citep{geirhos2020shortcut,altakrori2021topic}.

A key challenge is that style and content are often entangled in naturally collected data. Texts with different style labels may also differ in topics, entities, events, or domains, which allows a classifier to exploit content cues that are predictive of the label~\citep{altakrori2021topic}. Standard held-out evaluation may fail to diagnose this behavior: when the same content--style association is preserved across training and test splits, a shortcut-based model~\citep{geirhos2020shortcut,niven2019probing} can perform well while failing to learn style features that generalize across content.

Prior work has shown that NLP models can rely on unintended correlations in the data, including shortcut learning in general~\citep{geirhos2020shortcut,mccoy2019right,zhou2024explore} and topic-based shortcuts in style-related tasks~\citep{altakrori2021topic}. Probing studies have further examined what information is captured by learned representations~\citep{niven2019probing,conneau-etal-2018-cram}. Existing evaluations can reveal shortcut behavior, but they usually do not control the strength of the content shortcut. As a result, it remains unclear when a model moves from content cues to content-invariant style representation learning.

To address this gap, we study style classification under controlled content overlap. The evaluation is built from parallel texts, where the same content can be expressed in different styles, while the amount of content shared across style labels is systematically varied. We use an overlap parameter $\alpha = 1 - I(C;S)/H(S)$ to control
this variation, where $C$ is content identity and $S$ is the style
label. This parameter measures how much content is shared across
style classes, and is independent of the number of style classes. At $\alpha=0$, each style is associated with distinct content, so content alone can predict the style label. At $\alpha=1$, all styles share the same content, so content no longer provides label information. English Bible translations~\citep{carlson2018evaluating, christodouloupoulos2015massively} provide an aligned testbed for this setup, with shared content expressed across multiple stylistic variants. 

Our contributions are threefold.  First, we introduce an information-based measure of controlled
content overlap $\alpha$ that
quantifies the strength of content shortcuts independently of the
number of style classes. Second, we introduce a cross-overlap evaluation and show that matched accuracy can hide content shortcuts: low-overlap models perform well under matched conditions but degrade sharply when content cues are no longer predictive, whereas high-overlap training leads to more stable transfer. Third, we introduce a cross-style content retrieval probe showing that content identity becomes less recoverable as $\alpha$ increases and that this change happens gradually during training. 
\noindent\textbf{Code.} The code is available at \url{https://github.com/joeliuz6/content_overlap_eval}.

\section{Evaluation Setup}
\label{sec:method}

\subsection{Task and Data Structure}

We consider a parallel corpus where the same content (indexed by a \emph{chunk ID}~$c$) appears in $k$ different style versions. Each data point is a triple $(x, s, c)$: the text~$x$, the style label~$s \in \{1, \dots, k\}$, and the chunk~ID~$c$. A style classifier takes $x$ as input and predicts $s$. The content shortcut arises when different styles see different chunks at training time.

\subsection{Controlled Overlap Sampling}
\label{sec:overlap}

We define content overlap using an information-based parameter:
\begin{equation}
\alpha = 1 - \frac{I(C;S)}{H(S)},
\label{eq:alpha_def}
\end{equation}
where $C$ denotes chunk identity and $S$ denotes the style label.
This parameter measures how much style information is not explained by content identity.
When $\alpha=0$, content identity fully determines the style label; when $\alpha=1$, content identity gives no information about the style label.

We use a simple sampling to construct datasets with a target value of $\alpha$.
Given $k$ style versions and a pool of $C$ chunks available in all versions, we set the per-version chunk number to $n = \lfloor C/k \rfloor$.
Each version's chunks come from two pools:

\begin{itemize}
    \item \textbf{Shared pool} ($n_p = \lfloor n\alpha \rfloor$ chunks): chunks that are included in all versions.
    \item \textbf{Exclusive pool} ($n_r = n - n_p$ chunks): chunks that are assigned to only one version.
\end{itemize}

This sampling strategy realizes the target information-based overlap. Intuitively, shared chunks do not predict the style label, while exclusive chunks fully predict it. Under this sampling, the conditional entropy is $H(S \mid C)=\alpha \log_2 k$, and therefore $I(C;S)=(1-\alpha)\log_2 k$. Substituting this into Eq.~\ref{eq:alpha_def} recovers the target value of $\alpha$, independent of the number of style classes. A full derivation is provided in Appendix~\ref{app:mi_proof}.

\begin{table*}[t]
\centering
\small

\begin{tabular}{lcccccc}
\toprule
& \multicolumn{6}{c}{$\aeval$} \\
\cmidrule(lr){2-7}
$\atrain$ & 0.0 & 0.2 & 0.4 & 0.6 & 0.8 & 1.0 \\
\midrule
0.0 & \textbf{94.9{\scriptsize$\pm$0.2}} & 82.2{\scriptsize$\pm$0.5} & 63.8{\scriptsize$\pm$0.4} & 56.2{\scriptsize$\pm$0.5} & 55.1{\scriptsize$\pm$0.8} & 55.3{\scriptsize$\pm$0.6} \\
0.2 & 88.5{\scriptsize$\pm$0.6} & \textbf{94.7{\scriptsize$\pm$0.4}} & 81.5{\scriptsize$\pm$0.8} & 67.0{\scriptsize$\pm$1.3} & 64.6{\scriptsize$\pm$2.5} & 65.4{\scriptsize$\pm$1.5} \\
0.4 & 84.2{\scriptsize$\pm$0.8} & 89.9{\scriptsize$\pm$0.3} & \textbf{91.5{\scriptsize$\pm$0.6}} & 82.8{\scriptsize$\pm$1.5} & 74.4{\scriptsize$\pm$0.7} & 75.5{\scriptsize$\pm$1.7} \\
0.6 & 84.8{\scriptsize$\pm$1.4} & 86.2{\scriptsize$\pm$1.1} & 88.9{\scriptsize$\pm$2.0} & \textbf{89.5{\scriptsize$\pm$1.0}} & 79.8{\scriptsize$\pm$1.4} & 81.8{\scriptsize$\pm$1.5} \\
0.8 & 85.6{\scriptsize$\pm$0.9} & 86.0{\scriptsize$\pm$0.5} & 85.3{\scriptsize$\pm$1.9} & 85.9{\scriptsize$\pm$1.3} & \textbf{87.8{\scriptsize$\pm$0.4}} & 86.7{\scriptsize$\pm$0.9} \\
1.0 & 87.4{\scriptsize$\pm$0.4} & 88.3{\scriptsize$\pm$0.4} & 88.6{\scriptsize$\pm$2.2} & 88.0{\scriptsize$\pm$0.9} & 85.9{\scriptsize$\pm$0.1} & \textbf{86.7{\scriptsize$\pm$1.0}} \\
\bottomrule
\end{tabular}
\caption{Cross-evaluation \textbf{test accuracy}~(\%) for $k = 2$. Rows: $\atrain$; columns: $\aeval$. Bold: matched conditions. Results are reported as mean{\scriptsize$\pm$}std over three random seeds. Results for $k = 3,4,5$ are shown as Table~\ref{tab:cross_test_acc_full} in Appendix.}
\label{tab:cross_test_k2}
\end{table*}
\begin{figure*}[t]
\centering
\includegraphics[width=2.0\columnwidth]{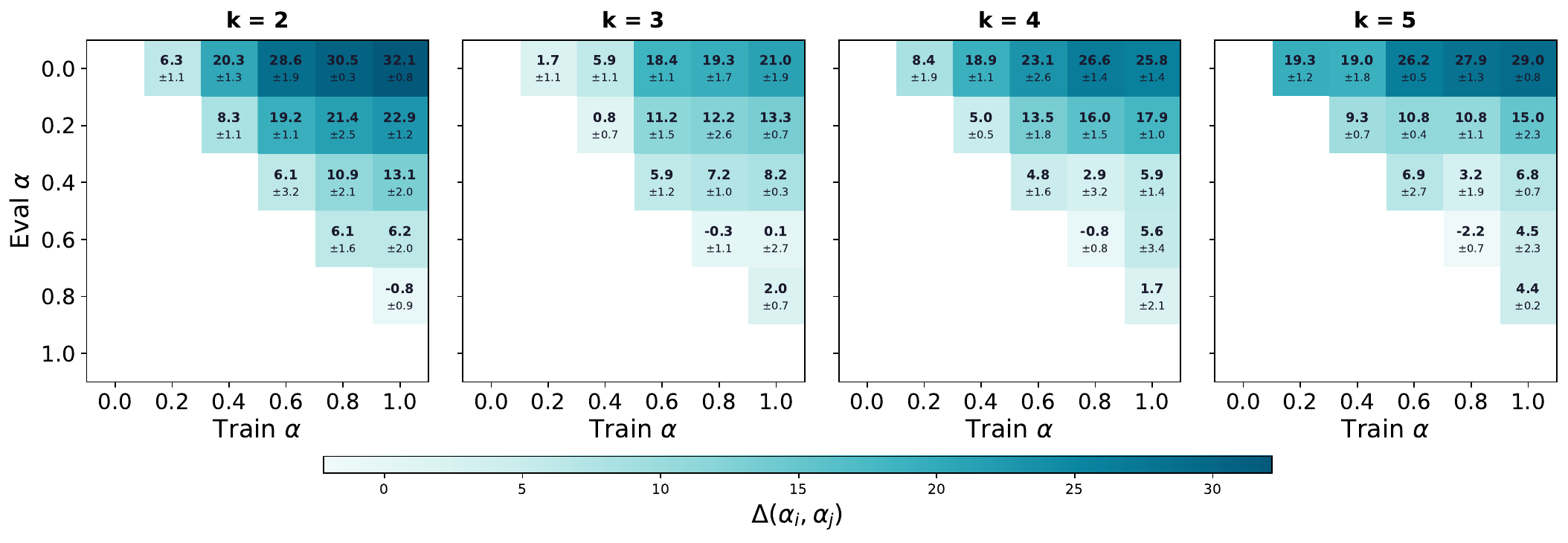}
\caption{
Higher-Overlap Advantage $\Delta_{\mathrm{HOA}}${\scriptsize$\pm$}std (Eq.~\ref{eq:hoa}) for all pairs of training and evaluation overlaps, for $k \in \{2,3,4,5\}$. 
Each cell shows the accuracy gain of the higher-overlap model relative to the lower-overlap model. }
\label{fig:upper_tri_test}
\end{figure*}

\subsection{Cross-Overlap Evaluation}
\label{sec:cross_eval}

We use cross-overlap evaluation to measure how well learned style features transfer when the content--style association changes. We train a RoBERTa-large~\citep{liu2019roberta} classifier with an MLP classification head for each training overlap level $\atrain$, fine-tuning all model parameters, and then evaluated on datasets with every overlap level $\aeval$.

Raw cross-overlap accuracy is not directly comparable across different values of $\aeval$, since evaluation overlap levels can differ in difficulty. We therefore compare models under same overlap shifts.
For $\alpha_i < \alpha_j$, we compare a model trained at $\alpha_j$ and tested at $\alpha_i$ against a model trained at $\alpha_i$ and tested at $\alpha_j$. This isolates the effect of training overlap, which we measure with the \emph{higher-overlap advantage}:
\begin{multline}
\Delta_{\mathrm{HOA}}(\alpha_i, \alpha_j) =
\text{Acc}(\atrain{=}\alpha_j,\; \aeval{=}\alpha_i) \\
- \text{Acc}(\atrain{=}\alpha_i,\; \aeval{=}\alpha_j),
\label{eq:hoa}
\end{multline}
for \(\alpha_i < \alpha_j\). A positive \(\Delta_{\mathrm{HOA}}(\alpha_i, \alpha_j)\) indicates that the model trained with higher overlap transfers better under the same shift. For example, $\Delta_{\mathrm{HOA}}(0.0, 1.0)$ compares the $\atrain{=}1.0$ model evaluated at $\aeval{=}0.0$ against the $\atrain{=}0.0$ model evaluated at $\aeval{=}1.0$; a positive value means the high-overlap model generalizes downward better than the low-overlap model generalizes upward.

\subsection{Cross-Style Content Retrieval Probe}
\label{sec:probe_method}

To measure how much content information is retained in the learned representation, we introduce a \emph{cross-style content retrieval probe}. Our style classifier uses RoBERTa-large with an MLP classification head. The \texttt{[CLS]} representation from RoBERTa is passed into the MLP head, and we use the intermediate representation $\mathbf{h}$ from this MLP head for probing. For each trained style classifier, we freeze the model and extract the intermediate representation $\mathbf{h}$. We then train lightweight linear projectors that map representations from two styles into a shared embedding space. The projectors are trained with a CLIP-style contrastive loss~\citep{radford2021learning}, where aligned pairs from the same chunk but different versions are pulled together, and non-aligned pairs are pushed apart. Thus, the probe measures whether content is linearly recoverable after a linear projection.

We evaluate content recoverability using top-1 bidirectional retrieval accuracy with cosine similarity as in the paper~\citep{radford2021learning}. Given a chunk representation from one style, the probe retrieves its nearest neighbor from another style; the prediction is correct if the retrieved text has the same chunk ID. High retrieval accuracy indicates that chunk identity remains recoverable from the representation. In addition to probing fully trained models, we apply the same probe at different epochs during classifier training, allowing us to track how content recoverability changes. Full details of the probe are provided in Appendix~\ref{content-matching}.

\section{Experimental Setup}
\label{sec:experiments}

\subsection{Data}
 
We use seven English Bible translations as our parallel corpus. Each translation is segmented into aligned chunks of consecutive verses. For classification, we use a subset of five translations, yielding style class settings $k \in \{2,3,4,5\}$. The other two translations are reserved exclusively for the unseen cross-style content retrieval probe. Full dataset statistics and processing details are in Appendix~\ref{app:data}.
 
\subsection{Implementation Details}
\label{sec:implementation}

We finetune all RoBERTa-large~\citep{liu2019roberta} parameters over six overlap levels, from $\alpha=0.0$ to $\alpha=1.0$ in increments of $0.2$. Main results are averaged over three random seeds. Full details are in Appendix~\ref{app:training}.

\section{Results}
\label{sec:results}

We present three complementary analyses: cross-overlap evaluation tests whether accuracy transfers when content--style associations change, the retrieval probe measures whether content remains recoverable, and training dynamics show when this content information is removed.
\subsection{High-Overlap Training Improves Cross-Overlap Style Transfer}
\label{sec:cross_eval_results}

Table~\ref{tab:cross_test_k2} shows the cross-overlap evaluation matrix for $k = 2$. The diagonal entries are high for all training overlaps, but matched accuracy alone can be misleading. For example, the model trained at \(\atrain=0.0\) reaches 94.9\% under matched evaluation, but drops to 55.3\% when evaluated at \(\aeval=1.0\). This shows that low-overlap training encourages content shortcuts: when content cues are removed at evaluation time, performance collapses. In contrast, the model trained at \(\atrain=1.0\) remains stable across all \(\aeval\) values (85--88\%). Since content is fully shared during training, content identity no longer predicts the style label, forcing the model to rely more on stylistic features. This pattern holds across different numbers of classes as shown in Table~\ref{tab:cross_test_acc_full} in Appendix.

\begin{figure}[t]
    \subfigure[Seen versions]{
        \includegraphics[width=0.47\columnwidth]{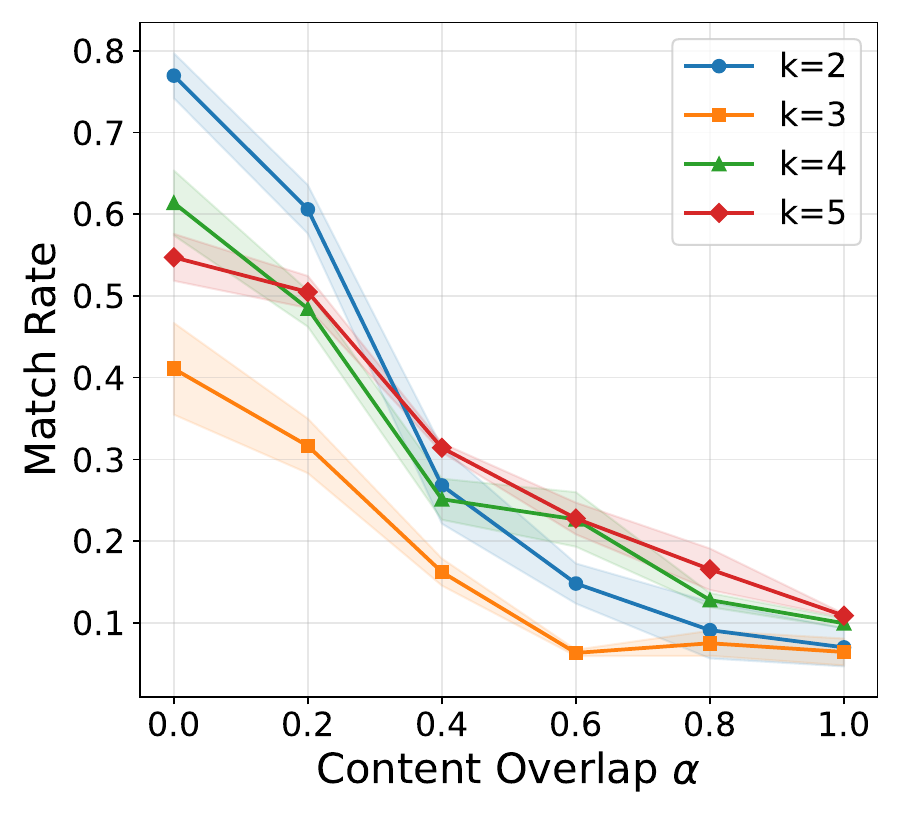}
    }
     \subfigure[Unseen versions]{
        \includegraphics[width=0.47\columnwidth]{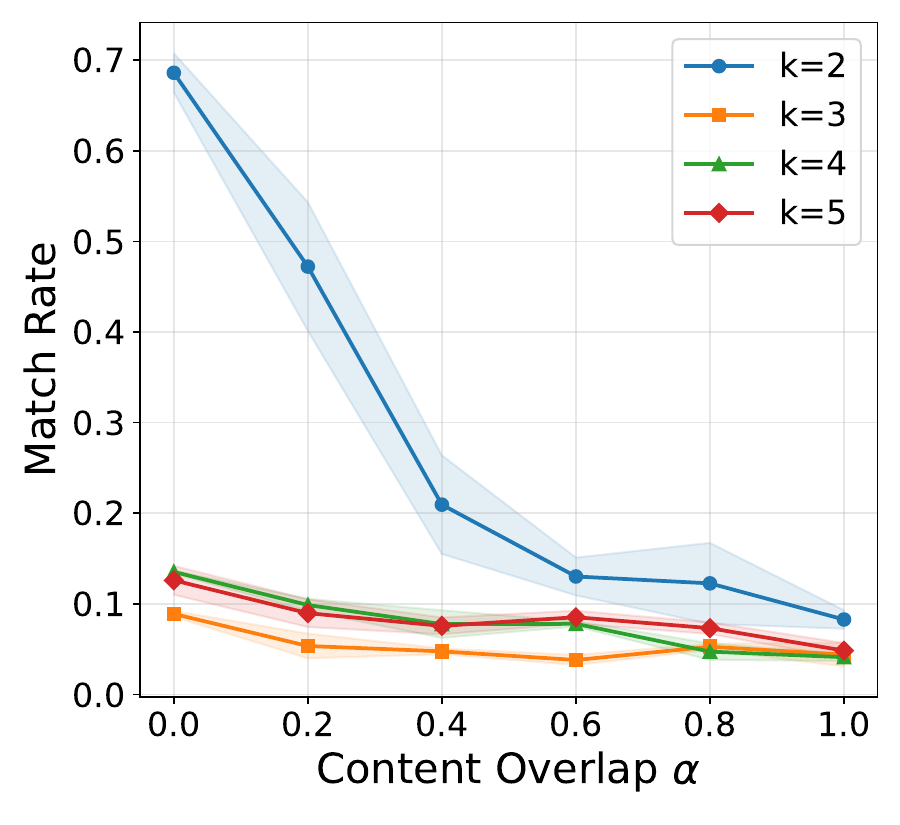}
    }
    \caption{Cross-style content retrieval probe accuracy vs.\ training overlap $\alpha$ for different $k$. Shaded regions indicate standard deviation across three random seeds.}
     \label{fig:match}
\end{figure}

Figure~\ref{fig:upper_tri_test} visualizes the high-overlap advantage in Eq.~\ref{eq:hoa} for each $k$.

Nearly all entries are positive, showing that higher-overlap training consistently produces representations that transfer better. This advantage is statistically reliable: a one-sided paired $t$-test over all off-diagonal pairs rejects the null hypothesis of no directional advantage ($p < 0.001$ for all $k$), with large effect sizes (Cohen's $d > 1.1$) and over 88\% of pairs favoring the higher-overlap model (Appendix~\ref{app:significance}).

\subsection{Content Decreases with Overlap}
\label{sec:matching}

The content retrieval probe measures how much content information remains in the learned representations. Figure~\ref{fig:match} reports top-1 retrieval accuracy across classifiers trained with different overlap levels $\alpha$. For the retrieval evaluation, we use the fully shared setting ($\alpha=1.0$). Thus, the training overlap of the style classifier varies from $\alpha=0.0$ to $\alpha=1.0$, while the retrieval probe is evaluated on shared-content pairs.

Retrieval accuracy decreases monotonically as $\alpha$ increases for all $k$. At $\alpha=0$, retrieval accuracy is high (above 0.7), showing that low-overlap models strongly encode content. At $\alpha=1$, it drops sharply (below 0.1), indicating that most content information has been removed.
The same trend holds on unseen versions, although seen versions retain slightly higher retrieval accuracy. Style classification accuracy remains high even at $\alpha=1$ (Table~\ref{tab:cross_test_acc_full} in Appendix), showing that the model shifts from encoding content to encoding style. These results support the information view: as $\alpha$ increases, content becomes less predictive of style, and the learned representation suppresses content features.

\begin{figure}[t]
    \subfigure[Seen versions]{
        \includegraphics[width=0.47\columnwidth]{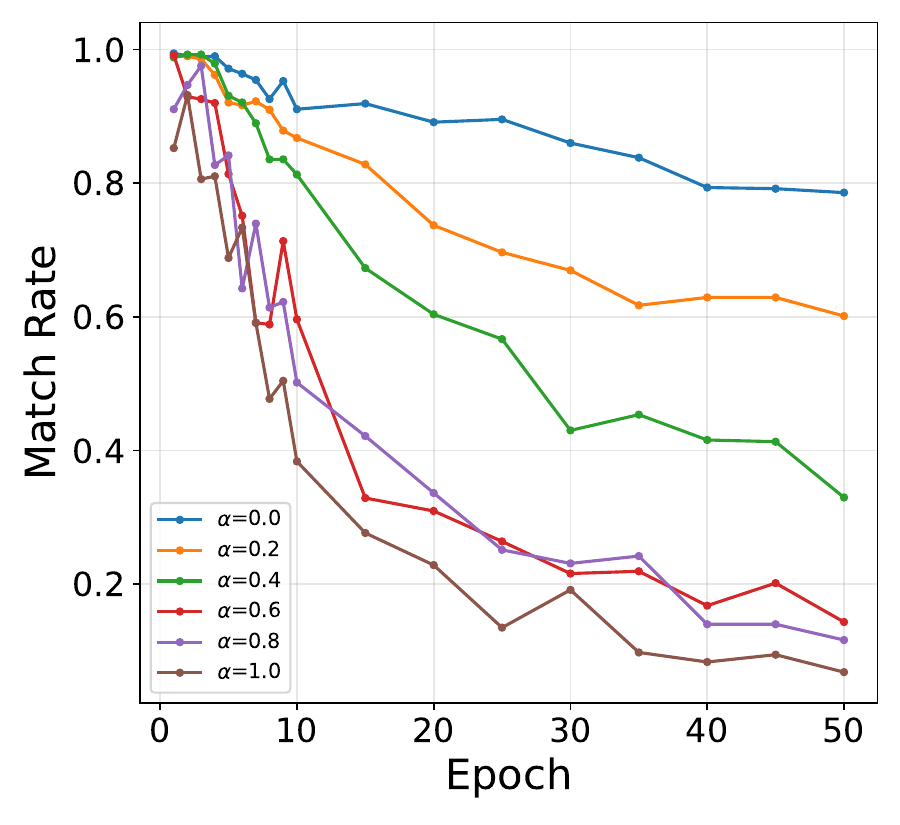}
    }
     \subfigure[Unseen versions]{
        \includegraphics[width=0.47\columnwidth]{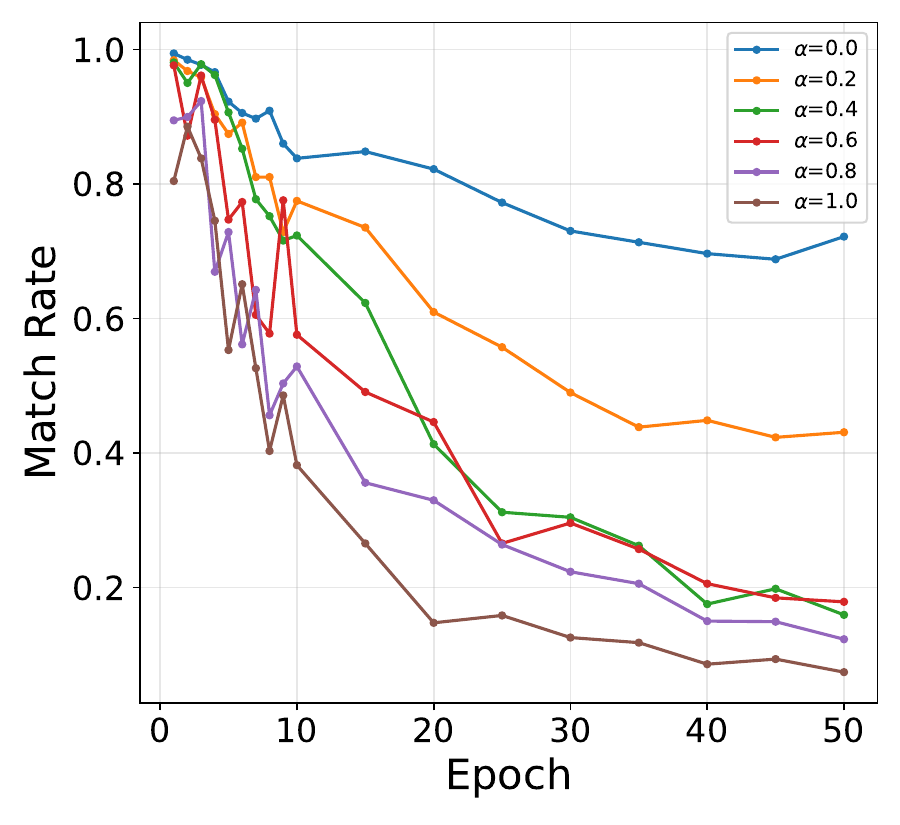}
    }
    \caption{Cross-style content retrieval probe accuracy during training for different $\atrain$ with $k = 2$.}
     \label{fig:dynamics}
\end{figure}
\subsection{Training Dynamics of Content Removal}
\label{sec:dynamics}


Figure~\ref{fig:dynamics} shows retrieval accuracy over training for different $\alpha$ values with $k=2$. At the beginning, all models have similarly high matching accuracy, reflecting the pretrained RoBERTa initialization. As training proceeds, the trajectories separate clearly.
High-overlap models remove content much faster: by epoch 20, the $\alpha=1.0$ model drops below 0.2, while the $\alpha=0$ model remains above 0.8. This indicates that low-overlap continues to preserve content because it is useful for predicting the style.

Across all $\alpha$, we observe the same pattern: larger $\alpha$ leads to faster and deeper content removal. Thus, content removal is gradual rather than sudden, and its speed is controlled by the overlap structure of the training data. Since style classification accuracy remains high, this process reflects a shift from content-based shortcuts toward style-based representations rather than representational collapse.

\section{Conclusion}
\label{sec:conclusion}

Our findings point to a practical diagnostic setup: when content and style co-vary in training data, standard held-out accuracy is unreliable as evidence of style learning. The overlap parameter $\alpha$ offers a straightforward way to verify this. We hope this encourages the community to treat content bias as a measurable, adjustable variable rather than an unquantified source of noise in style-related benchmarks.

\section*{Limitations}

\paragraph{Limited model coverage.}
We evaluate RoBERTa-large with an MLP classification head. 
Although this setting is sufficient for testing the controlled-overlap setting, future work should examine newer encoder models, decoder-only language models, and instruction-tuned models.

\paragraph{Limited dataset scope.}
We use English Bible translations because they provide aligned content across stylistic variants. 
However, this domain does not cover all forms of style variation, such as authorship, register, genre, or social media style. 
Testing additional aligned or semi-aligned corpora would strengthen the generality of our findings.

\paragraph{No semantic-distance modeling between chunks.}
Our setup treats chunk identity as discrete: chunks are either identical or different. 
It therefore does not capture semantic relatedness between different chunks. 
The mutual-information analysis should be interpreted as measuring exact content overlap, not broader semantic overlap.

\paragraph{Limited overlap schedule.}
We use six overlap levels from $\alpha=0.0$ to $\alpha=1.0$ in steps of 0.2. 
A finer grid could reveal more detailed transition patterns between content-based shortcuts and more content-invariant style representations.

\bibliography{latex/custom.bib}

\clearpage

\appendix

\section{Derivation of the Information-Based Overlap}
\label{app:mi_proof}

We provide the complete derivation for the deterministic sampling scheme ($p = \alpha$, $r = 1 - \alpha$).

\paragraph{Chunk structure.}
Under this scheme, every chunk falls into exactly one of two categories:
(1)~Shared chunks: there are $n_p = n\alpha$ such chunks, each appearing in all $k$ versions (multiplicity $m_c = k$).
(2)~Exclusive chunks: there are $k \cdot n_r = kn(1-\alpha)$ such chunks, each appearing in exactly one version (multiplicity $m_c = 1$).

\paragraph{Verification.}
Total rows: $n\alpha \cdot k + kn(1-\alpha) \cdot 1 = kn = N$.

\paragraph{Style entropy.}
Each version has exactly $n$ rows out of $N = kn$ total, so $P(s) = 1/k$ and $H(S) = \log_2 k$.

\paragraph{Conditional entropy.}
For a shared chunk ($m_c = k$): $P(s \mid c) = 1/k$, so $H(S \mid C = c) = \log_2 k$.
Each such chunk has weight $P(c) = k/(kn) = 1/n$.
For an exclusive chunk ($m_c = 1$): $H(S \mid C = c) = 0$.
Each such chunk has weight $P(c) = 1/(kn)$.
Summing:
\begin{align}
H(S \mid C)
&= \sum_{\text{shared}} \frac{k}{kn} \log_2 k
+ \sum_{\text{excl.}} \frac{1}{kn} \cdot 0
\notag \\
&= n\alpha \cdot \frac{1}{n} \cdot \log_2 k
= \alpha \log_2 k.
\label{eq:hsc}
\end{align}

\paragraph{Mutual information.}
$$I(C;S) = H(S) - H(S \mid C) = (1 - \alpha) \log_2 k.$$

\paragraph{Normalized residual.}
$$\frac{I(C;S)}{H(S)} = \frac{(1-\alpha)\log_2 k}{\log_2 k} = 1 - \alpha,$$
hence $1 - I(C;S)/H(S) = \alpha$.

\begin{table*}[t]
\centering
\small

\begin{tabular}{clcccccc}
\toprule
\multicolumn{2}{c}{} & \multicolumn{6}{c}{$\aeval$} \\
\cmidrule(lr){3-8}
$k$ & $\atrain$ & 0.0 & 0.2 & 0.4 & 0.6 & 0.8 & 1.0 \\
\midrule
\multirow{6}{*}{$2$} & 0.0 & \textbf{94.9{\scriptsize$\pm$0.2}} & 82.2{\scriptsize$\pm$0.5} & 63.8{\scriptsize$\pm$0.4} & 56.2{\scriptsize$\pm$0.5} & 55.1{\scriptsize$\pm$0.8} & 55.3{\scriptsize$\pm$0.6} \\
 & 0.2 & 88.5{\scriptsize$\pm$0.6} & \textbf{94.7{\scriptsize$\pm$0.4}} & 81.5{\scriptsize$\pm$0.8} & 67.0{\scriptsize$\pm$1.3} & 64.6{\scriptsize$\pm$2.5} & 65.4{\scriptsize$\pm$1.5} \\
 & 0.4 & 84.2{\scriptsize$\pm$0.8} & 89.9{\scriptsize$\pm$0.3} & \textbf{91.5{\scriptsize$\pm$0.6}} & 82.8{\scriptsize$\pm$1.5} & 74.4{\scriptsize$\pm$0.7} & 75.5{\scriptsize$\pm$1.7} \\
 & 0.6 & 84.8{\scriptsize$\pm$1.4} & 86.2{\scriptsize$\pm$1.1} & 88.9{\scriptsize$\pm$2.0} & \textbf{89.5{\scriptsize$\pm$1.0}} & 79.8{\scriptsize$\pm$1.4} & 81.8{\scriptsize$\pm$1.5} \\
 & 0.8 & 85.6{\scriptsize$\pm$0.9} & 86.0{\scriptsize$\pm$0.5} & 85.3{\scriptsize$\pm$1.9} & 85.9{\scriptsize$\pm$1.3} & \textbf{87.8{\scriptsize$\pm$0.4}} & 86.7{\scriptsize$\pm$0.9} \\
 & 1.0 & 87.4{\scriptsize$\pm$0.4} & 88.3{\scriptsize$\pm$0.4} & 88.6{\scriptsize$\pm$2.2} & 88.0{\scriptsize$\pm$0.9} & 85.9{\scriptsize$\pm$0.1} & \textbf{86.7{\scriptsize$\pm$1.0}} \\
\midrule
\multirow{6}{*}{$3$} & 0.0 & \textbf{96.5{\scriptsize$\pm$0.3}} & 90.0{\scriptsize$\pm$0.5} & 78.2{\scriptsize$\pm$1.1} & 68.5{\scriptsize$\pm$1.1} & 68.3{\scriptsize$\pm$1.3} & 68.8{\scriptsize$\pm$1.3} \\
 & 0.2 & 91.7{\scriptsize$\pm$0.9} & \textbf{95.8{\scriptsize$\pm$0.6}} & 89.6{\scriptsize$\pm$0.4} & 77.7{\scriptsize$\pm$0.3} & 76.0{\scriptsize$\pm$0.9} & 77.6{\scriptsize$\pm$0.4} \\
 & 0.4 & 84.2{\scriptsize$\pm$1.4} & 90.3{\scriptsize$\pm$0.4} & \textbf{94.4{\scriptsize$\pm$0.9}} & 85.0{\scriptsize$\pm$0.9} & 80.5{\scriptsize$\pm$1.2} & 81.8{\scriptsize$\pm$0.6} \\
 & 0.6 & 87.0{\scriptsize$\pm$0.0} & 88.9{\scriptsize$\pm$1.7} & 91.0{\scriptsize$\pm$0.7} & \textbf{90.2{\scriptsize$\pm$1.1}} & 86.7{\scriptsize$\pm$0.5} & 87.4{\scriptsize$\pm$0.8} \\
 & 0.8 & 87.6{\scriptsize$\pm$0.7} & 88.2{\scriptsize$\pm$1.8} & 87.8{\scriptsize$\pm$0.3} & 86.4{\scriptsize$\pm$0.8} & \textbf{91.0{\scriptsize$\pm$0.3}} & 87.9{\scriptsize$\pm$0.5} \\
 & 1.0 & 89.8{\scriptsize$\pm$0.9} & 90.9{\scriptsize$\pm$0.5} & 90.0{\scriptsize$\pm$0.6} & 87.5{\scriptsize$\pm$3.0} & 89.9{\scriptsize$\pm$0.3} & \textbf{89.3{\scriptsize$\pm$0.2}} \\
\midrule
\multirow{6}{*}{$4$} & 0.0 & \textbf{97.6{\scriptsize$\pm$0.2}} & 78.3{\scriptsize$\pm$1.5} & 62.2{\scriptsize$\pm$0.9} & 54.3{\scriptsize$\pm$0.8} & 54.9{\scriptsize$\pm$0.7} & 55.6{\scriptsize$\pm$0.5} \\
 & 0.2 & 86.8{\scriptsize$\pm$0.4} & \textbf{92.8{\scriptsize$\pm$0.2}} & 81.2{\scriptsize$\pm$0.3} & 67.4{\scriptsize$\pm$0.6} & 67.7{\scriptsize$\pm$1.0} & 68.2{\scriptsize$\pm$1.1} \\
 & 0.4 & 81.1{\scriptsize$\pm$1.3} & 86.3{\scriptsize$\pm$0.4} & \textbf{88.2{\scriptsize$\pm$0.6}} & 75.7{\scriptsize$\pm$1.9} & 76.2{\scriptsize$\pm$1.0} & 74.7{\scriptsize$\pm$0.6} \\
 & 0.6 & 77.5{\scriptsize$\pm$1.8} & 80.9{\scriptsize$\pm$1.4} & 80.5{\scriptsize$\pm$0.6} & \textbf{83.7{\scriptsize$\pm$1.1}} & 81.1{\scriptsize$\pm$1.4} & 75.4{\scriptsize$\pm$1.8} \\
 & 0.8 & 81.5{\scriptsize$\pm$0.8} & 83.7{\scriptsize$\pm$1.0} & 79.1{\scriptsize$\pm$2.2} & 80.3{\scriptsize$\pm$0.7} & \textbf{86.0{\scriptsize$\pm$1.3}} & 80.2{\scriptsize$\pm$0.9} \\
 & 1.0 & 81.4{\scriptsize$\pm$1.8} & 86.2{\scriptsize$\pm$0.3} & 80.6{\scriptsize$\pm$0.9} & 81.0{\scriptsize$\pm$1.6} & 81.9{\scriptsize$\pm$1.2} & \textbf{80.8{\scriptsize$\pm$1.1}} \\
\midrule
\multirow{6}{*}{$5$} & 0.0 & \textbf{96.0{\scriptsize$\pm$0.8}} & 68.1{\scriptsize$\pm$1.6} & 60.4{\scriptsize$\pm$1.0} & 50.9{\scriptsize$\pm$0.9} & 52.9{\scriptsize$\pm$0.9} & 52.1{\scriptsize$\pm$0.8} \\
 & 0.2 & 87.4{\scriptsize$\pm$1.1} & \textbf{88.0{\scriptsize$\pm$0.7}} & 71.8{\scriptsize$\pm$1.4} & 64.2{\scriptsize$\pm$0.9} & 67.1{\scriptsize$\pm$0.8} & 65.0{\scriptsize$\pm$0.8} \\
 & 0.4 & 79.4{\scriptsize$\pm$0.8} & 81.1{\scriptsize$\pm$1.4} & \textbf{85.4{\scriptsize$\pm$0.5}} & 70.9{\scriptsize$\pm$1.1} & 73.0{\scriptsize$\pm$0.9} & 69.4{\scriptsize$\pm$0.4} \\
 & 0.6 & 77.1{\scriptsize$\pm$0.5} & 75.0{\scriptsize$\pm$0.8} & 77.8{\scriptsize$\pm$1.9} & \textbf{81.9{\scriptsize$\pm$0.6}} & 79.0{\scriptsize$\pm$0.4} & 73.6{\scriptsize$\pm$0.6} \\
 & 0.8 & 80.7{\scriptsize$\pm$0.4} & 77.8{\scriptsize$\pm$1.8} & 76.2{\scriptsize$\pm$1.1} & 76.8{\scriptsize$\pm$0.7} & \textbf{83.5{\scriptsize$\pm$0.7}} & 76.3{\scriptsize$\pm$1.2} \\
 & 1.0 & 81.2{\scriptsize$\pm$0.7} & 80.0{\scriptsize$\pm$2.0} & 76.2{\scriptsize$\pm$0.8} & 78.0{\scriptsize$\pm$1.8} & 80.7{\scriptsize$\pm$1.4} & \textbf{79.2{\scriptsize$\pm$1.7}} \\
\bottomrule
\end{tabular}
\caption{Cross-evaluation \textbf{test accuracy}~(\%) for $k \in \{2,3,4,5\}$. Results are reported as mean{\scriptsize$\pm$}std over 3 random seeds.}
\label{tab:cross_test_acc_full}
\end{table*}

\section{Cross-Style Content Retrieval Probe details}
\label{content-matching}
 
The cross-style content retrieval probe measures how much content information is retained in the style classifier's representation $\mathbf{h}$. For all retrieval-probe evaluations, we construct the probe dataset under the fully shared setting ($\alpha=1.0$), so that aligned chunks are available across the two versions used for retrieval.
The overlap value varied in the main analysis refers to the training overlap of the style classifier. We use two seen versions and two unseen version translations for any value of $k$.
 
Given a trained and frozen style classifier, we extract $\mathbf{h} \in \mathbb{R}^{256}$ for every chunk in both seen and unseen verison translations.
Two linear projectors $W^a, W^b \in \mathbb{R}^{64 \times 256}$ (one per translation) map $\mathbf{h}$ into a shared 64-dimensional space, trained with a symmetric contrastive loss:
\begin{equation}
\begin{aligned}
\mathcal{L}
= -\frac{1}{2N}\sum_{i=1}^{N} \bigg[
& \log
\frac{
e^{s(\mathbf{z}_i^a,\mathbf{z}_i^b)/\tau}
}{
\sum_j e^{s(\mathbf{z}_i^a,\mathbf{z}_j^b)/\tau}
}
\\
&+
\log
\frac{
e^{s(\mathbf{z}_i^b,\mathbf{z}_i^a)/\tau}
}{
\sum_j e^{s(\mathbf{z}_i^b,\mathbf{z}_j^a)/\tau}
}
\bigg],
\end{aligned}
\label{eq:contrastive}
\end{equation}
where $\mathbf{z}_i^a = W^a \mathbf{h}_i^a$ and $\mathbf{z}_i^b = W^b \mathbf{h}_i^b$ are projected embeddings of an aligned pair (same chunk in two different translations), $s(\cdot,\cdot)$ is cosine similarity, and $\tau = 0.07$ is the temperature.
We report top-1 bidirectional matching accuracy: high accuracy means the representation retains content information; low accuracy means content has been stripped away. For the training dynamics experiments (\S\ref{sec:dynamics}), the same probe is trained from scratch at regular epoch intervals using the classifier checkpoint at that point.

\section{Dataset}
\label{app:data}

\begin{table*}[h]
\centering
\small
\begin{tabular}{llc}
\toprule
\textbf{Translation} & \textbf{Abbr.} & \textbf{Role} \\
\midrule
American Standard Version & ASV & Classification + Seen Probe \\
Darby Bible Translation & DBY & Classification + Seen Probe \\
World English Bible & WEB & Classification \\
King James Version & KJV & Classification \\
Webster's Bible Translation & WBT & Classification \\
\midrule
English Revised Version & ERV & Unseen Probe \\
World Messianic Bible & WMB & Unseen Probe \\
\bottomrule
\end{tabular}
\caption{Bible translations versions used in this study. The top five are used for style classification ($k \in \{2,3,4,5\}$ in listed order); the bottom two are reserved for the unseen cross-style content retrieval probe.}
\label{tab:translations}
\end{table*}

We selected the seven English translations from the English Bible versions available through eBible.org, which provides multiple downloadable English translations including ASV, DBY, ERV, KJV, WEB, WBT, and WMB. We first applied an alignment-coverage filter: a candidate translation was retained only if it covered enough of the shared verse/chapter inventory to support strict parallel chunking across versions. This step removed translations with substantial missing books, chapters, or verse ranges, since such gaps would make chunk-level alignment unreliable and would introduce systematic missingness unrelated to style.

We then filtered candidates using two complementary diagnostic criteria. First, we excluded versions that were too easily distinguishable from the others because of obvious surface artifacts, such as highly archaic or simplified language, distinctive orthography, paraphrastic expansions, naming conventions, or formatting/editorial conventions. Such versions could allow a classifier to rely on superficial cues rather than on the subtler relationship between content and style. Second, we removed translations that were extremely similar to another candidate version, since near-duplicate translations would make the classification setting artificially dependent on minute textual differences and would reduce the diversity of stylistic variation in the corpus. This near-duplication is common in Bible translation corpora because many English versions are revisions, editions, or light modernizations of earlier translations rather than fully independent translations. As a result, two versions may share extensive verse-level wording even when they are distributed as separate translations. We therefore aimed to remove both outliers and near-duplicates, retaining translations that were similar enough to support controlled aligned comparison but different enough to provide meaningful stylistic variation.

Table~\ref{tab:translations} shows the seven English Bible translations used in this study. For each $k \in \{2, 3, 4, 5\}$, we use a fixed nested subset of five classification translations (ASV, DBY, WEB, KJV, WBT): $k=2$ uses the first two, $k=3$ the first three, and so on.
The remaining two translations (ERV, WMB) are reserved exclusively for the unseen cross-style content retrieval probe. For the seen version probe, we use ASV and DBY. Each translation is segmented into chunks of $L = 5$ consecutive verses, producing aligned chunk~IDs across all seven versions.
Chunks where all translations produce identical text are removed, since they carry no stylistic signal.

Data is split by chunk~ID into train chunks, validation, and test sets with an 80/10/10 ratio, so that no content chunk appears in more than one split.
For each split, overlap-controlled subsets are constructed independently using the same target $\alpha \in \{0.0, 0.2, 0.4, 0.6, 0.8, 1.0\}$.
For a given $\alpha$ and $k$ versions, each version receives $n = \lfloor C_{\text{split}} / k \rfloor$ chunks, of which $n_p = \lfloor n\alpha \rfloor$ are shared across all versions and the remaining $n_r = n - n_p$ are exclusive to that version. Exclusive chunks are assigned sequentially; shared chunks are drawn randomly from the remaining pool.
 
\section{Training Details}
\label{app:training}

\paragraph{Style classifier.}
The classifier is based on RoBERTa-Large~\citep{liu2019roberta} with an MLP classification head:
(1)~a dense layer $\mathbf{h} = \tanh(W_1 \cdot \mathbf{h}_{\texttt{[CLS]}} + b_1)$, where $\mathbf{h} \in \mathbb{R}^{256}$;
(2)~a linear classifier $\hat{y} = W_2 \cdot \mathbf{h} + b_2$, where $\hat{y} \in \mathbb{R}^{k}$.
The model is trained end-to-end with cross entropy loss using AdamW with learning rate $2 \times 10^{-5}$, weight decay $0.01$, linear warmup for 10\% of total steps, batch size 32 with gradient accumulation over 4 steps (effective batch size 128), and 50 epochs. The maximum input length is 512 tokens.

\paragraph{Content retrieval probe.}
The projectors are linear layers mapping from the 256-dimensional representation $\mathbf{h}$ to a 64-dimensional embedding space (one projector per version). They are trained with AdamW (learning rate $2 \times 10^{-3}$, weight decay $10^{-3}$, $\beta = (0.9, 0.98)$), batch size 256, 100 epochs, cosine annealing, and a temperature of $\tau = 0.07$ for the contrastive loss. The best checkpoint is selected by validation matching accuracy.
For the training dynamics experiments (\S\ref{sec:dynamics}), the same probe is trained from scratch at regular epoch intervals using the classifier checkpoint at that point.

All classification and probe results are averaged over three random seeds.

\section{Statistical Significance of the High-Overlap Advantage}
\label{app:significance}
 
To test whether the directional pattern in the cross-evaluation matrices is statistically significant, we pool all off-diagonal pairs $(\alpha_i, \alpha_j)$ with $\alpha_i < \alpha_j$ and compute the high-overlap advantage $\Delta(\alpha_i, \alpha_j)$ (Eq.~\ref{eq:hoa}) for each pair under each random seed.
This yields $\binom{6}{2} \times 3 = 45$ paired differences per $k$.
We test the null hypothesis $H_0{:}\; \mathbb{E}[\Delta] \leq 0$ (i.e., higher overlap confers no directional advantage) using a one-sided $t$-test and a Wilcoxon signed-rank test.
 
Table~\ref{tab:significance} reports the results on the test set.
Both tests reject $H_0$ at $p < 0.001$ for every $k$.
The effect is large and consistent: Cohen's $d$ exceeds 1.1 in all conditions, and over 88\% of individual pair differences are positive, confirming that the benefit of higher-overlap training is systematic rather than driven by a few extreme pairs.
 
\begin{table}[t]
\centering
\small

\begin{tabular}{ccccc}
\toprule
$k$ & Mean $\Delta$ (\%) & Cohen's $d$ & $\%\,{>}\,0$ & $p$ ($t$-test) \\
\midrule
2 & 15.4 & 1.52 & 93.3 & $<$0.001 \\
3 &  8.5 & 1.19 & 88.9 & $<$0.001 \\
4 & 11.7 & 1.29 & 88.9 & $<$0.001 \\
5 & 12.7 & 1.34 & 93.3 & $<$0.001 \\
\bottomrule
\end{tabular}
\caption{Statistical significance of the high-overlap advantage $\Delta$ on the test set. For each $k$, 45 paired differences are pooled across 15 off-diagonal pairs and 3 seeds. Both the $t$-test and Wilcoxon test yield $p < 0.001$ in all conditions.}
\label{tab:significance}
\end{table}

\section{Additional Representation Geometry Analysis}
\label{app:spectrum}

\begin{figure}[t]
\centering
\includegraphics[width=1.0\columnwidth]{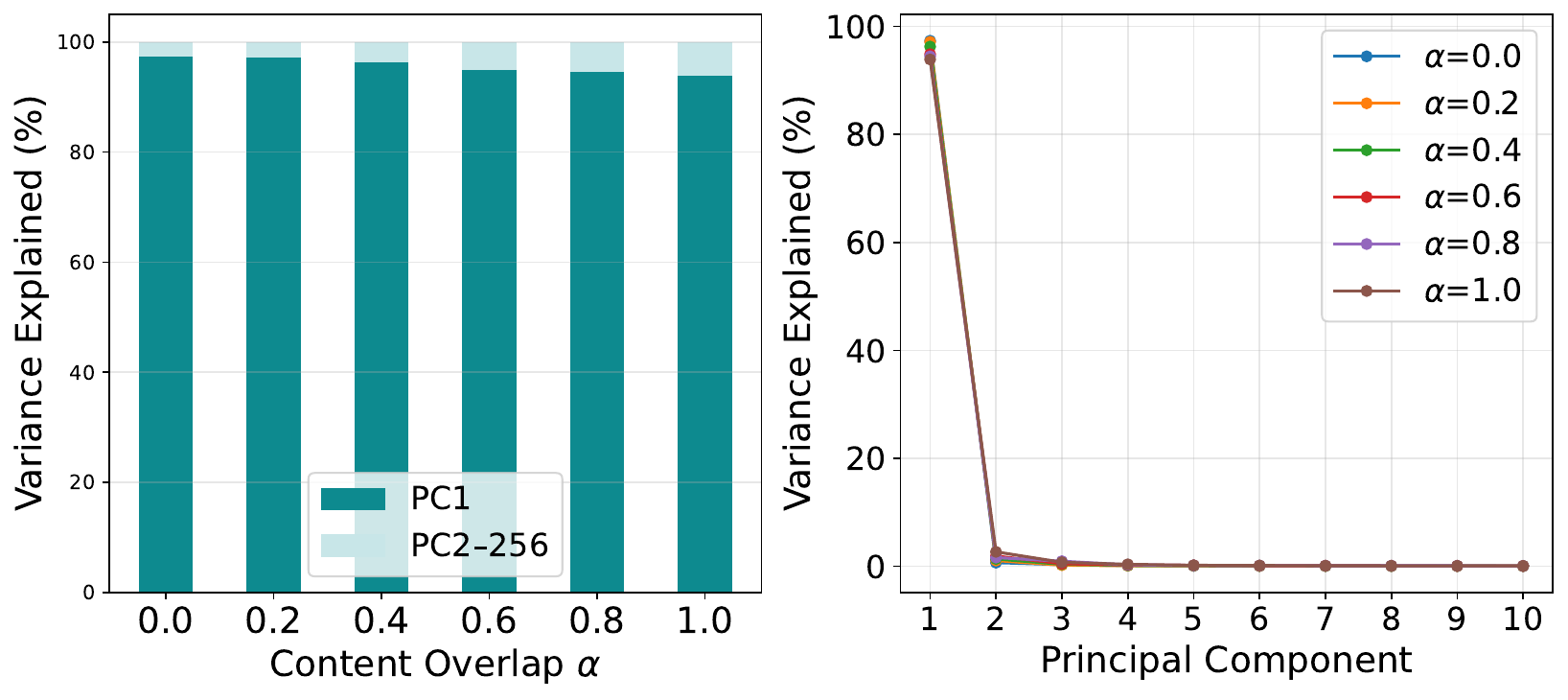}
\caption{
Spectrum analysis of the 256-dimensional pre-logit representations for $k=2$.
Left: variance explained by PC1 and the remaining components across different training overlaps.
Right: variance explained by the top 10 principal components.
Across all overlap levels, the spectrum is dominated by PC1, suggesting that the classifier representation is highly anisotropic.
}
\label{fig:spectrum}
\end{figure}

\begin{table}[t]
\centering
\small

\begin{tabular}{ccc}
\toprule
$\atrain$ & Effective Rank & Participation Ratio \\
\midrule
0.0 & 1.240 & 1.055 \\
0.2 & 1.247 & 1.059 \\
0.4 & 1.320 & 1.078 \\
0.6 & 1.442 & 1.110 \\
0.8 & 1.485 & 1.120 \\
1.0 & 1.489 & 1.133 \\
\bottomrule
\end{tabular}
\caption{
Effective rank and participation ratio of the mean-centered representation covariance matrix on the test set for $k=2$.
Both metrics are close to 1 across all overlap levels, indicating that most variance is concentrated in a single dominant direction.
}
\label{tab:geometry_test}
\end{table}

We include an additional analysis of the global geometry of the learned representation.
For each model trained with $k=2$, we extract the 256-dimensional pre-logit representation $\mathbf{h}$ on the test set.
Before computing the covariance matrix, we mean-center the representations.
Let $\lambda_1 \geq \lambda_2 \geq \cdots \geq \lambda_d$ be the eigenvalues of this covariance matrix, with $d=256$.

We summarize the spectrum using two standard measures.
The \emph{effective rank}~\citep{roy2007effective} is defined as $\exp(H)$, where
$H = -\sum_i \hat{\lambda}_i \log \hat{\lambda}_i$
and $\hat{\lambda}_i = \lambda_i / \sum_j \lambda_j$.
The \emph{participation ratio} is defined as
$(\sum_i \lambda_i)^2 / \sum_i \lambda_i^2$.
Both quantities are close to 1 when variance is concentrated in a single direction, and approach $d$ when variance is spread uniformly across dimensions.

Figure~\ref{fig:spectrum} shows that the spectrum is dominated by the first principal component across all overlap levels.
PC1 explains over 94\% of the variance, while the remaining components together account for less than 6\%.
Table~\ref{tab:geometry_test} shows the same pattern numerically: both effective rank and participation ratio remain close to 1 for all values of $\alpha$.

These results should be interpreted as a coarse geometry check rather than a direct measure of style or content information.
The strong dominance of PC1 likely reflects the anisotropy of the representation and the low-dimensional structure induced by the binary classification head.
Importantly, this global spectrum changes only slightly with $\alpha$, while the cross-style content retrieval probe shows large changes in content recoverability.
Thus, content information can become less recoverable even when coarse spectral statistics of the representation remain similar.

\section{Artifact Use and Licenses}
\label{app:artifact}

We use publicly available English Bible translations from eBible.org and the RoBERTa-large model. Before release, we will include the license information for each translation
and ensure that our redistribution of processed metadata follows the corresponding license terms. RoBERTa-large is released under the MIT License. Our use of these artifacts is consistent with their intended purposes: the Bible corpus was created as a multilingual parallel resource for NLP research, and RoBERTa-large is a general-purpose pretrained model designed for fine-tuning on downstream tasks. The dataset consists of publicly available religious texts and does not contain personally identifiable information or offensive content; no additional anonymization was required.

\end{document}